\documentclass[10pt,twocolumn,letterpaper]{article}

\usepackage{times}
\usepackage{epsfig}
\usepackage{graphicx}
\usepackage{amsmath}
\usepackage{amssymb}
\usepackage{epstopdf}
\usepackage[ruled,lined]{algorithm2e}
\usepackage{multirow}

\usepackage[pagebackref=true,breaklinks=true,letterpaper=true,colorlinks,bookmarks=false]{hyperref}

\newcommand{\minesubsec}[1]{\vspace{0.3cm} \noindent{\bf #1~}}

\newcommand{\BQ}{\begin{equation}}
\newcommand{\EQ}{\end{equation}}

\DeclareMathOperator*{\argmin}{argmin}
\DeclareMathOperator*{\argmax}{argmax}

\begin{document}
\title{Best-Buddies Tracking}

\author{Shaul Oron\\
{\tt\small shauloro@post.tau.ac.il}
\and
Denis Suhanov\\
{\tt\small denissuh@post.tau.ac.il}\\
\\
Tel Aviv University\\
\and
Shai Avidan\\
{\tt\small avidan@eng.tau.ac.il}
}

\maketitle

\begin{abstract}	
Best-Buddies Tracking (BBT) applies the Best-Buddies Similarity measure (BBS) to the problem of model-free online tracking. BBS was introduced as a similarity measure between two point sets and was shown to be very effective for template matching. Originally, BBS was designed to work with point sets of equal size, and we propose a modification that lets it handle point sets of different size. The modified BBS is better suited to handle scale changes in the template size, as well as support a variable number of template images. We embed the modified BBS in a particle filter framework and obtain good results on a number of standard benchmarks. 
\end{abstract}	
\section{Introduction}
Online tracking plays an important role in many computer vision applications such as autonomous driving, surveillance systems and human-computer-interface, to name a few. Tracking is a challenging task because of changes in view-point, illumination, and non rigid deformations of the object to be tracked. A tracker must strike a delicate balance between adapting to legitimate changes in the appearance of an object and adapting to background clutter thus causing drift.

These sources of variability require a representation that is invariant to them or the development of similarity measures that can handle them. In its most basic form, online tracking boils down to template matching, where the goal is to find a given template in the current image. This requires the definition of a similarity measure, such as minimizing the intensity difference between the template and the target. To complicate things, it is common to represent the position of an object with a bounding box. The bounding box often include some background pixels. These pixels are outliers that might confuse the matching function and cause drift.

We address these problems by using a new (dis)similarity measure called the Best-Buddies Similarity or BBS \cite{DekelEtAl15}. It is used within a particle filtering framework to produce what we call the Best-Buddies Tracker or BBT.

The template and candidate image regions are mapped to two point sets in some high dimensional space where BBS takes place. The mapping simply involves breaking the image region into small patches and creating vectors (i.e., high dimensional points) consisting of their pixel value and relative $xy$ position. BBS counts the number of Best-Buddies Pairs (BBPs) - pairs of points in source and target sets, where each point is the nearest neighbor of the other. This simple measure is quite robust to geometric deformations and considerable amount of outliers, making it an ideal candidate for tracking. 

Applying BBS to tracking requires computing BBS between sets of points with different size. For example, in case the target and candidate have different scales, or if multiple templates are used. 
However, the original BBS formulation does not handle scale change properly. Specifically, our analysis shows that BBS scores increase if one set is made larger while the other is kept fixed. We analyze this phenomenon both theoretically and empirically, and suggest several ways to address the problem. 

In addition to the scientific benefit, solving this issue is crucial for applying BBS to tracking. Otherwise, for example, comparing BBS scores of candidates of different scale (and hence point sets of different size) would be biased towards larger candidates.

Because BBS is a statistical measure, we show that it is enough to sample equal size point sets from the input point sets, even if the input point sets are of different size. A nice side benefit of this sampling strategy is that we can increase computational efficiency by sampling small size point sets. This is especially important in tracking where run time considerations are important.

In \cite{DekelEtAl15} BBS was used for template matching and was computed exhaustively over an image using a sliding window. In BBT we propose incorporating BBS into a particle filtering framework. This removes the need to perform an exhaustive search and allows us to account for scale changes as well. Additional features of BBT include, online confidence estimation using a forward-backward consistency check, and use of a tracker ensemble for more reliable scale estimation. 

Finally, we would like BBS to reason about object appearance changes over time. To this end, we leverage the fact we are working with point sets, and propose a ``bag-of-points'' approach, augmenting together points from multiple templates, captured at different times. By doing so, we obtain a better non-parametric representation of the underlying appearance model of the object as it changes over time, leading to more reliable matching.  

We perform extensive tests evaluating the performance of BBT. To this end, we use three commonly used tracking benchmarks, and compared our performance to many recently published methods. Overall, BBT demonstrates good initial performance, with significant improvement achieved through better ensemble fusion.
 
To summarize, the contribution of our work is two fold: (i) We address the problem of computing BBS with unbalanced point sets. We analyze the problem both theoretically and empirically, and propose two effective solutions to the problem. Solving this problem is crucial for applying BBS to tracking. (ii) We propose a novel tracking framework termed BBT that uses BBS as a similarity measure. We perform extensive experiments evaluating its performance and comparing them to other recently published methods showing promising initial results. 
\section{Related Work}
One can categorize a tracking algorithm as either taking a discriminative or generative approach to the problem. Within each category, one can use different representations to work with. On top of that, recent work demonstrated the benefits of using multiple trackers in parallel.

Discriminative trackers treat tracking as a binary classification problem where the goal is to develop a classifier that will separate the object from the background \cite{Avidan07,GrabnerLB08,HareST11,KalalMM12,henriques2015tracking,DSST2015,zhang2014meem}.

Generative trackers, on the other hand, build a model of the object and try to minimize reconstruction error when searching for the object in the next frame \cite{BlackJ98,RossLLY08,LiuYHMGK10}.

Within each approach one should determine the representation with which to work. Methods proposed in recent years often use multiple templates for a better representation of the time-varying appearance of the object. These templates are then used in various ways for computing the similarity. Some methods might use the raw pixel values. For example, using the templates as a dictionary and solving a sparse optimization problem \cite{SCM12,Zhang_2015_CVPR}. Treating the templates as a bag of patches and then using a patch matching based similarity measure \cite{OGT2014}. Other methods take a discriminative approach treating the similarity as a form of classification problem \cite{HongWMPT:ECCV2014,Liu_2015_CVPR}, or some combination of the two \cite{HareST11,KalalMM12,TGPR2014}.

Of course, one can mix and match different types of trackers. An early attempt was made by \cite{JepsonFE03} that combine a short term with a long term tracker. More recently, \cite{KwonL11} proposed a tracking sampling framework where new trackers are sampled in each new frame. Recently, the MUSTer tracker \cite{MUSTer2015} combined a short term tracker with a long term memory module to achieve excellent results. More general approaches for fusing results from an ensemble of trackers have also been proposed \cite{bailer2014superior,leichter2006general,wang2014ensemble}. In most such approaches the trackers themselves are treated as ``black-boxes'' and only their output signals are considered. 

Cardinal to any tracking algorithm is the need to determine the similarity measure between the appearance of an object and some candidate hypothesis. Such a measure must be defined for any type of tracker, be it a particle filtering (PF) approach \cite{SCM12,Mei2015}, a gradient decent based method \cite{Oron2014ELK}, exhaustive search over a region of interest \cite{HenriquesC0B15,HongWMPT:ECCV2014} or some other control schemes \cite{OGT2014,MUSTer2015}. Here we focus on the suitability of the BBS measure to the task.
\section{BBS with Unbalanced Set Sizes}
\label{sec: BBS}

%
%

We briefly review BBS \cite{DekelEtAl15} first. Then we discuss the problem of computing BBS when the point sets have different size. Solving this problem is crucial for proper visual tracking because the target and candidate point sets might have different size, either because of scale difference, or use of multiple templates.

\subsection{BBS for Template Matching}
BBS is a similarity measure between two point sets. So given a template and a target (in the form of two rectangular image region) we must first convert them to point sets. We do this by breaking each region into distinct patches of size $k \times k$ (we use $k=3$ in our experiments). Each such patch is represented by a vector of size $k^2d+2$ that is the concatenation of its $k^2$ pixels each with $d$ color channels and the $xy$ location of the central pixel, relative to the region's coordinate system. That is, each $k \times k$ patch is a point in a $\mathbb{R}^{k^2d+2}$ space. The template and the target are now represented as sets of points in this space. We are now ready to formally define BBS \cite{DekelEtAl15}:

BBS measures the similarity between two sets of points $P\!=\!\{p_i\}_{i=1}^N$ and $Q\!=\!\{q_i\}_{i=1}^M $, where  $p_i, q_i \in \mathbb{R}^n$.  The BBS is the fraction of \emph{Best-Buddies Pairs} (BBPs)  between the two sets. Specifically, a pair of points $\{ p_i \in P, q_j\in Q \}$ is a BBP if $p_i$ is the nearest neighbor of $q_j$  in the set $Q$, and vice versa. Formally,
{\small \BQ  \label{eq: bb1}
bb(p_i,q_j,P,Q) = \left\{
\begin{array}{ll}
	1 & {\text{NN}(p_i,Q)=q_j \land \text{NN}(q_j,P)=p_i}\\
	0 & \text{otherwise}
\end{array}
\right.\vspace{-0.02cm}\EQ} where, ${\text{NN}(p_i,Q)\!=\!\argmin\limits_{q\in Q}d(p_i,q)}$, and $d(p_i,q)$ is some distance measure. The BBS between the point sets $P$ and $Q$ is given by: \vspace{-0.04cm}
\BQ \label{eq: BB} \text{BBS}(P,Q) = \frac{1}{\min\{\text{M,N}\}}\cdot \sum_{i=1}^{N}\sum_{j=1}^M bb(p_i,q_j,P,Q).
\EQ

\subsection{BBS with Uneven Sets}\label{sec:bbs_claim}
BBS counts the number of Best-Buddy Pairs and then normalizes by the number of points in the \emph{smallest} set. As discussed in \cite{DekelEtAl15}, BBS is a statistical property of the data. It is governed by the underlying density functions from which the data was sampled. Moreover, it is shown in \cite{BBS_arxiv} that BBS converges to the Chi-Square ($\chi^2$) distance between distributions when the set sizes are sufficiently large.  

In general, BBS is well defined, and can be computed for uneven point sets. Unfortunately, in such cases BBS becomes biased. Specifically, when one set has fixed size, and we increase the size of the other set, the probability for finding BBP increases. However, the normalization factor is kept constant, thus making the final BBS score higher. 

To see this, consider the case where $P$ and $Q$ are drawn i.i.d from some underlying multivariate, $C^1$, distribution functions $f_P(p)$ and $f_Q(q)$. We keep the size of set $P$ fixed, and check what happens to the BBS score when $M = |Q|\rightarrow +\infty$.

\minesubsec{Claim:} $\underset{M\rightarrow+\infty}{lim}BBS(P,Q)\rightarrow 1$\\
This means that the BBS score goes to one when the size of set $Q$ goes to infinity (and the size of $P$ is fixed).

Intuitively, every point $p_i \in P$ has some region around it such that if a point $q \in Q$ is in this region then they are a BPP. As $Q$ becomes larger the probability that no point in $Q$ falls within this region gets smaller (for any point in $P$). This means that eventually all the points in $P$ will be BBP, and since we normalize by the size of the smaller set which is $P$, then the BBS score will goto one. For a more rigorous proof of this claim see the Appendix. 

\subsection{Making BBS Work With Uneven Sets}\label{sec:BBS sampling}

Uneven point sets affects the BBS score, and we propose two solutions to this problem. Key to both solutions is the understanding that if somehow these sets were made equal then the BBS computation will be unbiased.

Our first solution is clustering. Specifically, cluster the larger set such that it has the same size as the smaller set. This of course makes the sets equal in size but poses several problems. First, BBS is now computed relative to cluster centers which are not actual data points, but rather some form of averaged information. Second, clustering may, in some cases, change the underlying distribution of the data. Third, BBS disregards cluster weights which hold information regarding the underlying distribution. Finally, clustering adds additional computational load.

\begin{figure}[h!]	
	\centering
		\begin{tabular}{ccc}
			\includegraphics[width = 0.135\textwidth]{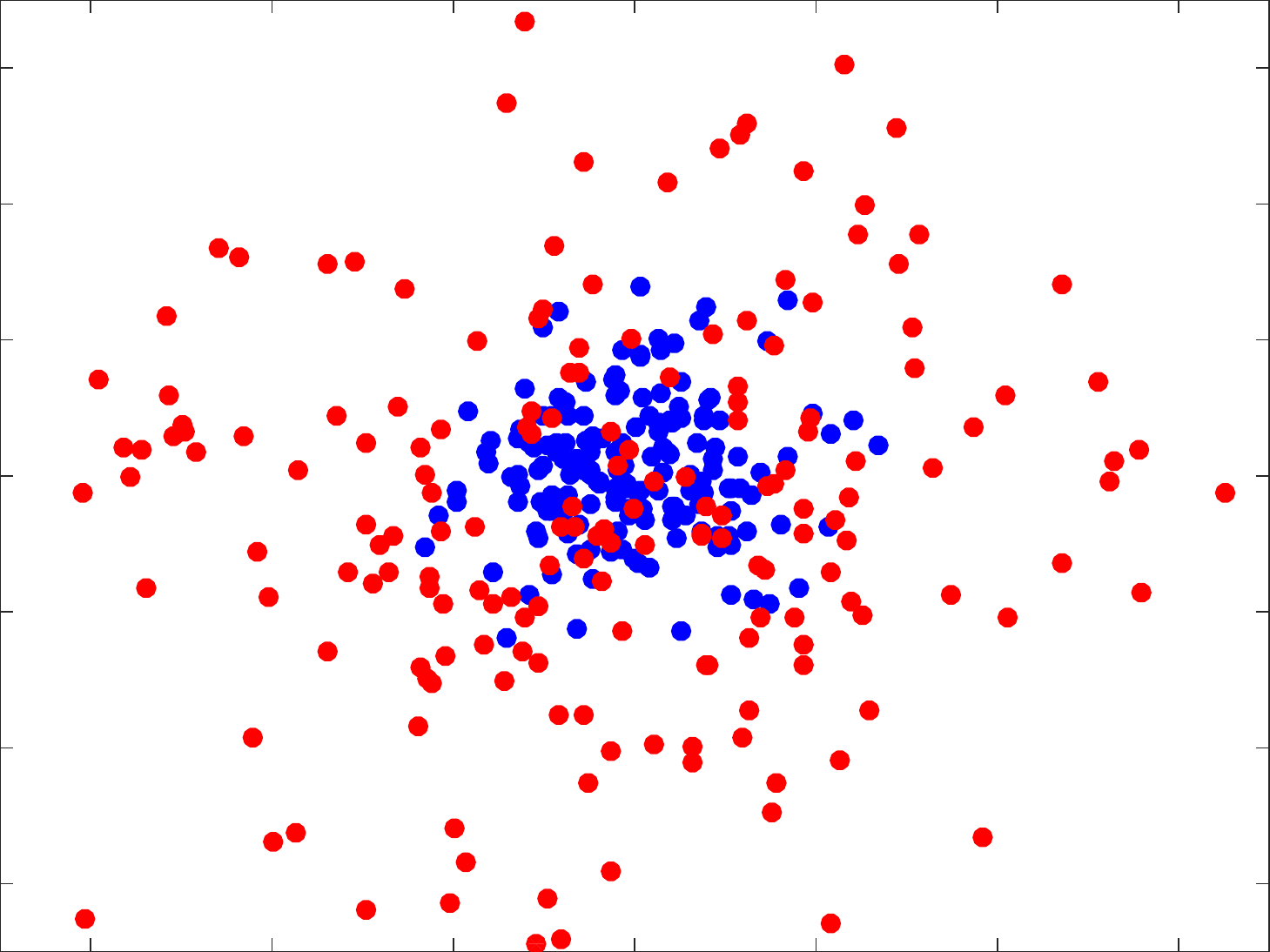} & 
			\includegraphics[width = 0.135\textwidth]{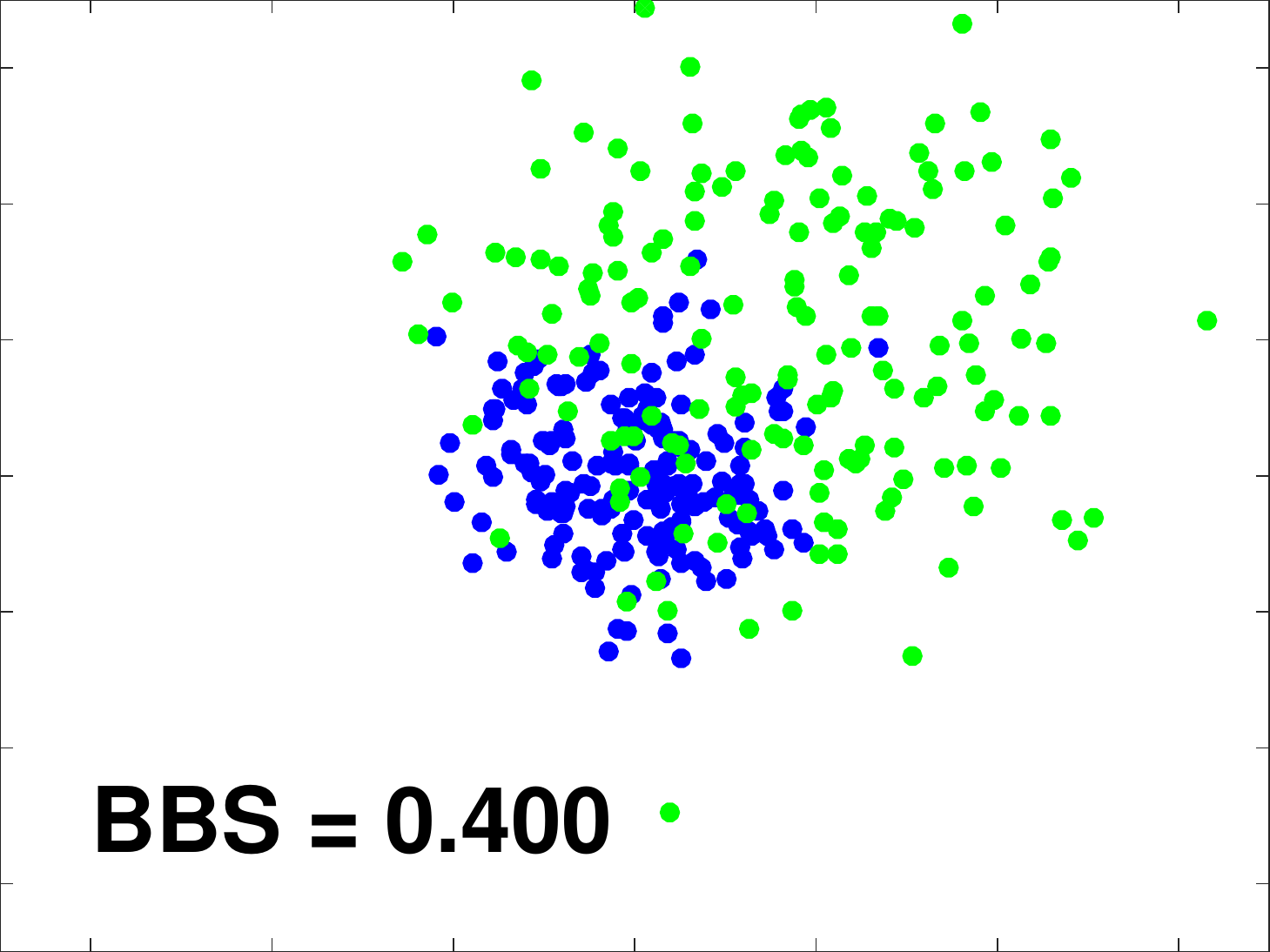} & 
			\includegraphics[width = 0.135\textwidth]{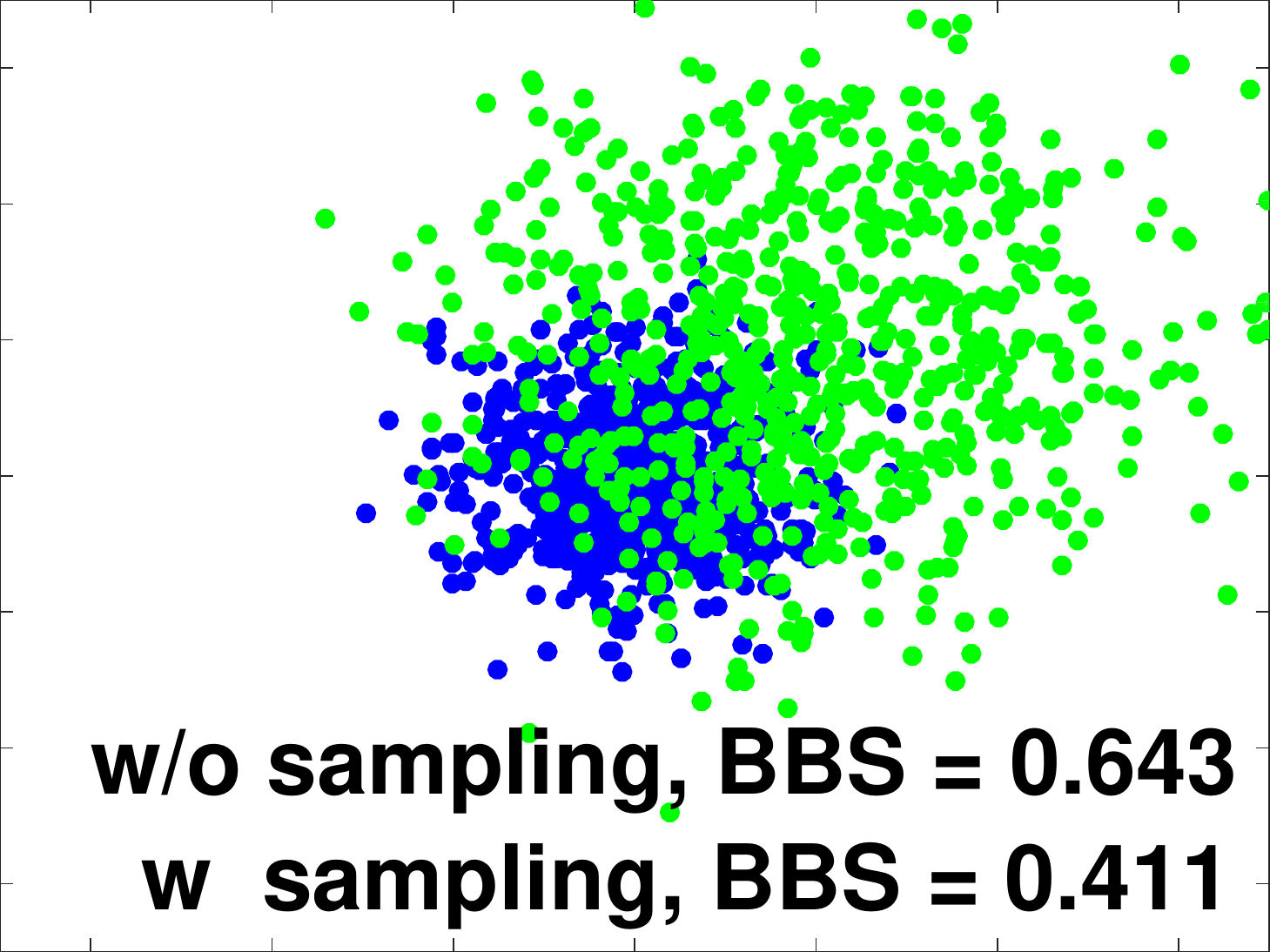}\\
			(a) & (b) & (c)\\
			\end{tabular}
 	\caption{{\bf 2D point sets:}. An example of point sets used in our synthetic experiment. Points are drawn from GMMs, which have the same ``Foreground'' Gaussian (Blue points) and a different ``Background'' Gaussian (Red/Green points). (a) Set $P$. (b) Set $Q$ s.t.  $|Q|=|P|$. (c) Set $Q$ s.t. $|Q|>|P|$. The BBS score between the even sets is shown in (b). BBS between the uneven sets, w/wo sampling is shown in (c). 	See text for more details.}
 	\label{fig: 2d points} \vspace{-0.3cm}
\end{figure}

These problems brings us to our second solution. Using random sampling instead of clustering. In this case we uniformly sample $K \leq \min(|P|,|Q|)$ points from both sets. Similar to Monte Carlo or particle filtering approaches, more points will be sampled from dense areas, where there is a higher probability of finding points. This solution alleviates the problems associated with clustering. Specifically, sampled points correctly reflect the true underlying distribution of the data, lifting the cluster weight problem. Unlike clustering, the sampled points are actual data points and not averaged cluster centers. No weights are needed, and the only pre-processing is the sampling itself. 

Finally, we note that a nice side product of the proposed techniques is the fact that they reduce the problem size, and are therefore expected to accelerate BBS computation. 

\minesubsec{Synthetic experiment:} The following synthetic experiment will illustrate the problem of computing BBS with uneven sets and demonstrate the effectiveness of our proposed solutions.
\begin{figure}[h!]	
	\centering
		\begin{tabular}{c}
			\includegraphics[width = 0.5\textwidth]{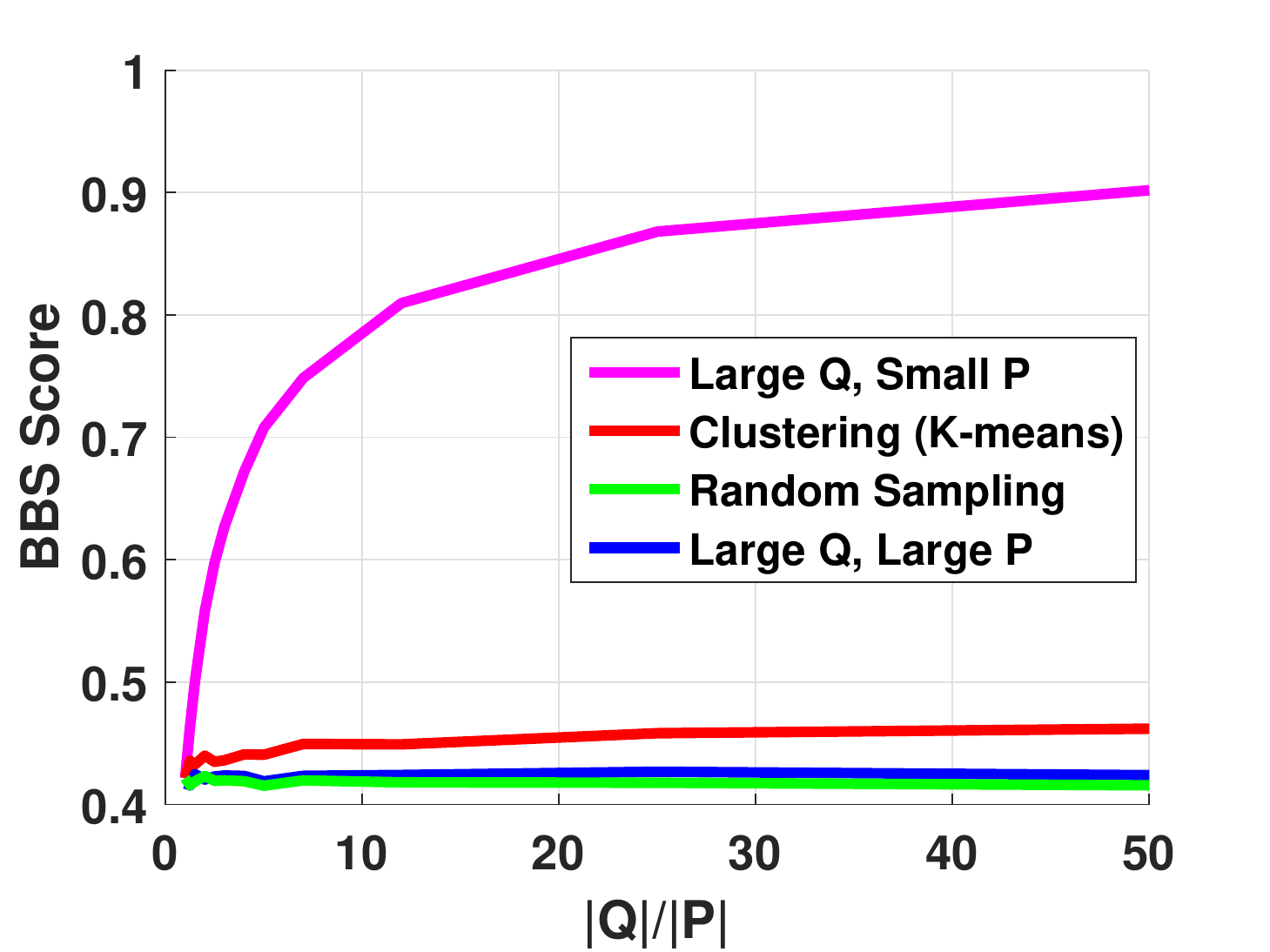} \\	
			\end{tabular}
 	\caption{{\bf BBS score for different set size ratios:} The number of point in $P$ is fixed while the size of $Q$ is increased. Increasing $Q$ makes it more likely for points in $P$ to find BBP resulting in higher BBS scores (Magenta). Using clustering (Red) and Random Sampling (Green) alleviate the score bias. BBS is invariant to set size when both sets are equally large (Blue).See text for more details}
 	\label{fig: BBS bias} \vspace{-0.3cm}
\end{figure}

Point sets $P$ and $Q$ consists of 2D points, drawn from underlying Gaussian-mixture-models (GMM), $f_P,f_Q$. 
Each GMM consists of two Gaussians. One is considered the ``Foreground'' Gaussian and the other the ``Background'' Gaussian. The ``Foreground'' Gaussian is the same for both GMMs while the ``Background'' one is different. An example showing point sets drawn using these distributions is shown in Figure \ref{fig: 2d points}. 

Ideally we want the BBS score of point sets $P$ and $Q$ to be invariant to the size of the point sets. Figure~\ref{fig: 2d points} reveals that this is not the case. The BBS score of $P$ and $Q$ is higher when the size of $Q$ grows.

Figure \ref{fig: BBS bias} shows the effect of our proposed solutions. Computing BBS using all points in $Q$ (Magenta curve) results in increasing BBS score as $|Q|$ increases. If both sets are increased equally the BBS score remains constant (Blue curve). Using clustering (Red curve) and random sampling (Green curve) eliminates the BBS bias and provides scores very close to this baseline.  Note how random sampling provides a better approximation to the correct BBS score compared with clustering.

\begin{figure} [h!]
	\centering		
			\includegraphics[width = 0.5\textwidth]{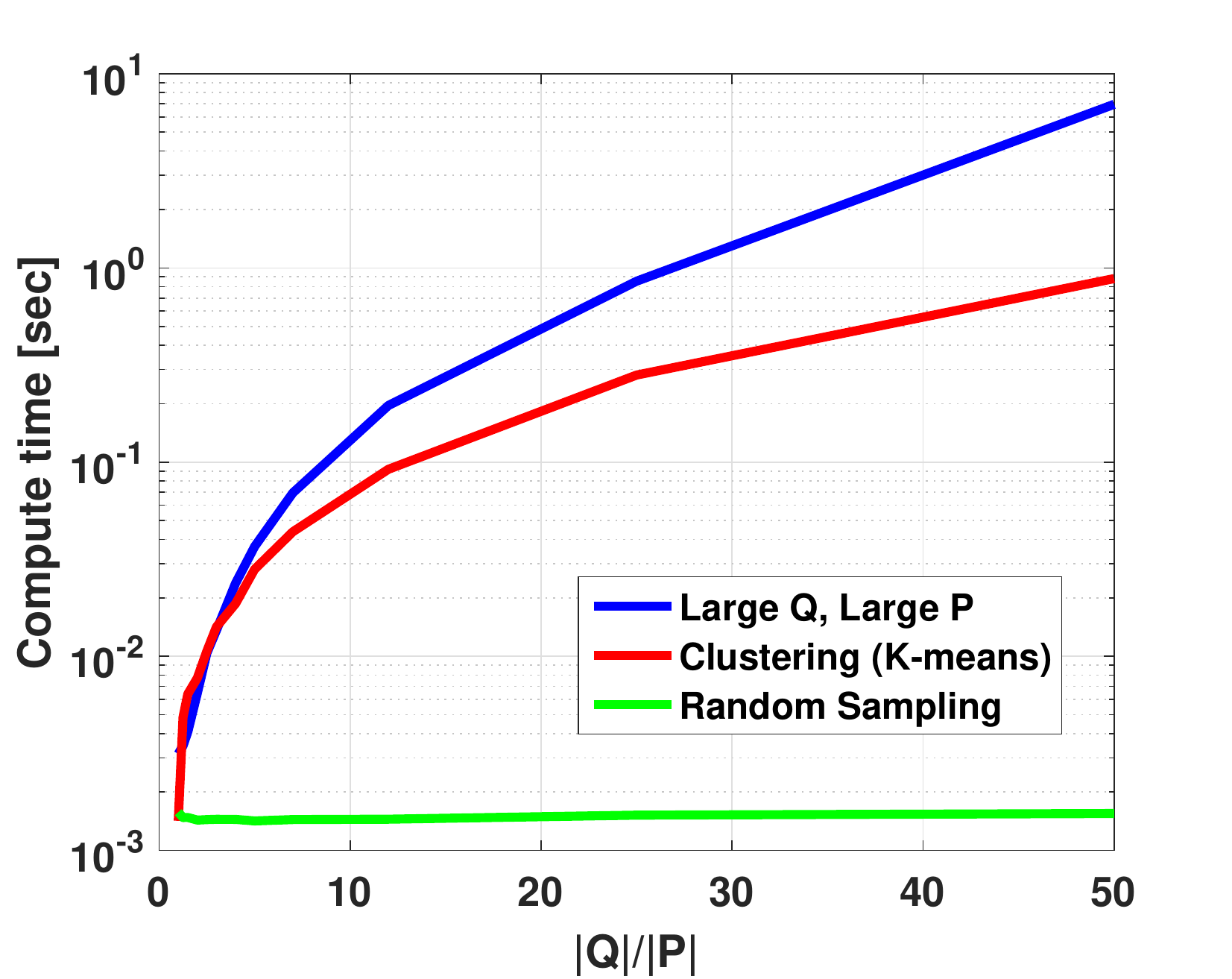}			
 	\caption{{\bf BBS computation time:} Computation time increases when both sets are made larger (Blue). Using clustering (Red) lowers the overall runtime but does not result in a constant processing time due to increased clustering compute time. Random sampling (Green) on the other hand does run in constant time.}
 	\label{fig: bbs time} \vspace{-0.3cm}
\end{figure}

Using either clustering or random sampling reduces the number of points used for the BBS computation and by doing so accelerate it. Figure \ref{fig: bbs time} shows average processing time of BBS as measured in our synthetic experiment. Note how using random sampling (Green) results in constant time processing. This is a desirable property for many application and specifically for tracking. Unfortunately, this is not the case for clustering (Red) which takes more time to compute as the size of $Q$ increases.  
\section{Best Buddies Tracker}
\label{sec:BBT}
In \cite{DekelEtAl15} BBS was successfully applied to template matching. It was shown to be robust to outliers and was able to account for template deformations. These properties make it appealing to use BBS as a similarity measure for visual tracking. Moreover, the data on which BBS was successfully evaluate in \cite{DekelEtAl15} was pairs of frames with a wide temporal baseline taken from a tracking benchmark (although it was not evaluated for tracking in that work).

In the original work, BBS was exhaustively applied to a query image using a sliding window. Despite the efficient computation scheme, proposed by the authors, this is still a computationally demanding process. Additionally, using the sliding window does not account for scale changes which are an inherent part of visual tracking. 

Due to these limitations, the proposed Best Buddies Tracker (BBT) uses a particle filtering framework. Each particle state represents a bounding box in the image ($X=[x,y,w,h]$). This way we can handle both scale and position changes. In addition, we avoid exhaustive computation. Our particle filtering framework closely follow the CONDENSATION algorithm\cite{IsardB98}, and  uses BBS to infer the observation likelihood of the particles.  

Computing BBS requires building a full distance matrix between all pairs of points in the template and candidate point sets. To this end we use the same distance measure as in the original BBS work. Specifically,  
\begin{equation} \label{eq:dist} \textsl{\textsl{}}
 d(p_i,q_j) = ||p_i^{(A)}-q_j^{(A)}||^2_2+\lambda||p_i^{(L)}-q_j^{(L)}||^2_2
\end{equation}
where superscript $A$ denotes a points appearance descriptor, which in our case are the $k^2d$ color channel values, and superscript $L$ denotes a points $xy$ location. We set $\lambda=2$ as in the original BBS work.

We deploy the random sampling approach, presented in Section \ref{sec:BBS sampling}, in order to avoid BBS bias due to uneven sets size. We expect this to also accelerate BBS computations. 

In order to better handle object appearance changes over time we use multiple templates, as will be explained next.
In addition, we add a forward-backward module to the particle filter. This module verifies that tracking from the current frame to some previous reference frame indeed lands at the position of the object at that frame. The output of the forward-backward module is a confidence score that is then used to determine whether to update the template set or not. This technique fits nicely with the BBS spirit that advocates the use of best buddies both in space and in time. 

Finally, in order to better handle scale changes and support both small scale changes as well as large abrupt changes we use an ensemble of several BBT trackers each with a different scale parameter setting. 

Detailed information on all the tracker components is provided next.

\minesubsec{Using Multiple Templates} 
In visual tracking objects may undergo a wide range of deformation as a result of in/out-of plane rotation, illumination changes, occlusions, articulation and more. In such cases using a static appearance of the object, e.g. the template from the first frame, will almost surely lead to drift. One of the common approaches for handling object appearance change over time is using multiple templates.  
In our case, BBT holds a template buffer comprised of $L$ templates chosen based of our forward-backward confidence score as will be explained next. For each input frame a subset of $l<L$ evenly spaced templates are taken from the buffer and used for the BBS computation as follows. 

\begin{figure*}
	\centering
		\begin{tabular}{c}
			\includegraphics[width = 1.0\textwidth]{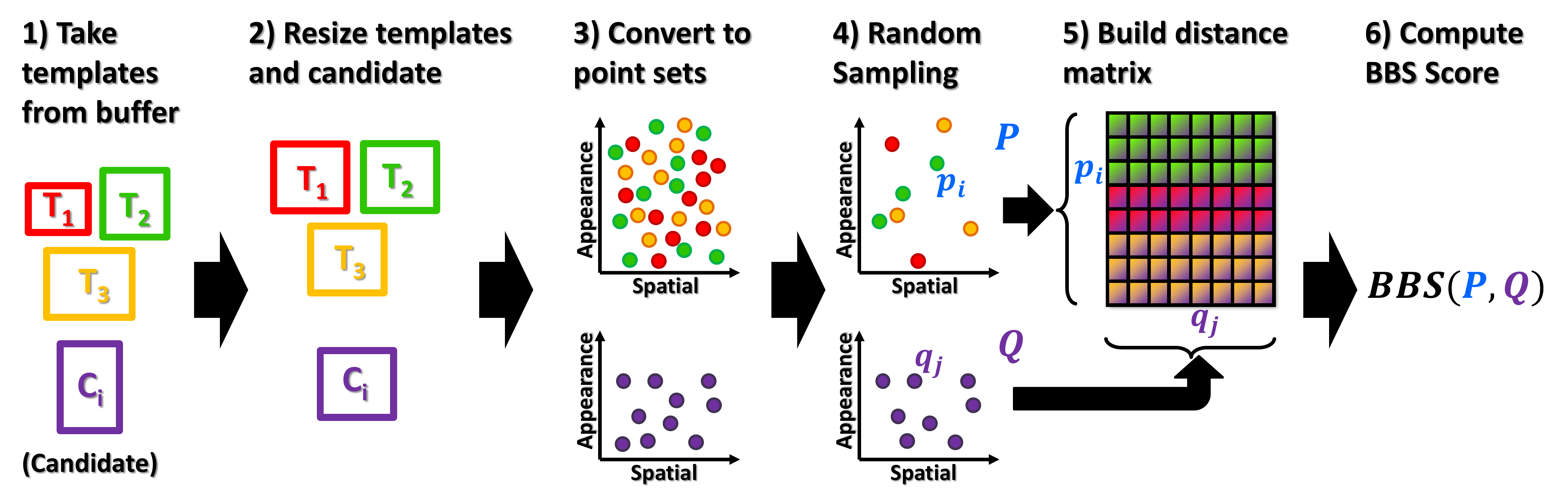} \\	
			\end{tabular}
 	\caption{{\bf BBS with Multiple Templates:} Multiple templates are taken from the template buffer (1), resized (2), and converted in to one big ``bag-of-points'' representing the objects appearance (3). Similarly the candidate window is also converted to a point set in a similar manner. Both sets are then randomly sampled (4) in order to ensure the BBS score will not be biased. Next a distance matrix is built (5) and BBS is computed (6). Since BBS is a statistical measure using a ``bag-of-points'' provides a better proxy for the underlying appearance of the object. See text for more details.}
 	\label{fig: multi-template} \vspace{-0.3cm}
\end{figure*}

Each one of these templates is resized to the average particle size. This is done in order to ensure that the spatial information stored in the $xy$ data of the feature space can be correctly leveraged. All points from all templates are then embedded in our spatial-appearance space. This essentially generates a ``bag'' of weakly localized feature points. Given some candidate window, we embed the points from that candidate in the same spatial-appearance space. In order to ensure unbiased BBS computation we randomly sample $K\leq\min(|P|,|Q|)$ points from both candidate and template point sets and only then compute the BBS. This entire process is illustrated in Figure \ref{fig: multi-template}.

We note that using using this approach means that a candidate point can find a BBP in any one of the templates used, allowing BBT to account for global as well as local deformations affecting only specific regions of the object. The reason using this ``bag-of-points'' representation makes sense is that BBS is a statistical property of the data. It accounts for the probability that points where drawn from the same underlying appearance model rather than that points are actual physical correspondences. By putting points from multiple templates in our ``bag'' we effectively obtain a better non-parametric representation of the objects underlying appearance.

\minesubsec{Forward-Backward Consistency}
Inspired by \cite{Kalal2010}, we use a forward-backward consistency check that produces a 
confidence score which estimates how well the tracker is locked on. This process is summarized in Algorithm \ref{alg:FWBK}. 

We begin by measuring the BBS score between the target in the current frame, $X^{cur}$, and candidates in a reference frame, which is some previous frame at which confidence was high. Candidates are taken on a grid around the target position at that frame, $X^{ref}$.

The confidence score is taken as the intersection over union between the state of the highest scoring candidate and $X^{ref}$. 
\begin{algorithm}[htb!]
\LinesNumbered
\small
\KwIn{\\
-Target appearance in current frame $T^{cur}=imcrop(I^{cur},X^{cur})$\\
-Reference frame $I^{ref}$\\
-State at reference frame $X^{ref}$
}
\KwOut{Confidence score $conf$}
$P\leftarrow$ convert $T^{cur}$ to point set \\
\For{states $X_i$ on grid around $X^{ref}$}{	
  $T^{ref}_i\leftarrow imcrop(I^{ref},X_i)$\\
	$Q\leftarrow$ convert $T^{ref}_i$ to point set\\
	Compute $bbs_i = BBS(P,Q)$\\
}
$X_{best} = \underset{X}{\argmax } \quad bbs_i$\\
$conf = \frac{X_{best}\cap X^{ref}}{X_{best}\cup X^{ref}}$\\
\caption{Forward-Backward Consistency Check}
\label{alg:FWBK}
\end{algorithm} 

\minesubsec{Updating Template and Reference Frame}
A new target template $T^t$ is added to the template buffer only if we were able to continue tracking, from that frame, with high confidence ($conf\ge\gamma_1$), for at least $f_1$ consecutive frames, and no other template was added in the last $f_1$ frames. That is, $T^t$ will be added to the template buffer at time $t+f_1$, only if, 
\begin{equation}\label{eq:t-update}
		conf^i\ge\gamma_1\; \forall\; i=\{t,t+1,...,t+f_1\}
\end{equation}
and no other template was buffered in the last $f_1$ frames.\\
Templates are buffered in a first-in-first-out (FIFO) manner with the only exception being $T^0$ which is never removed. For computational reasons, only $l$ equally spaced templates are used for the actual template matching at each frame.

Choosing reference frames is done in a similar process as updating templates. That is, frame $I^t$ will be considered a reference frame only at time $t+f_2$ and only if,
\begin{equation}\label{eq:ref-update}
		conf^i\ge\gamma_2\; \forall\; i=\{t,t+1,...,t+f_2\}
\end{equation}

The entire tracking flow for a frame streaming in is summarized in Algorithm \ref{alg:BBT}.
\begin{algorithm}[htb!]
\LinesNumbered
\small
\KwIn{New frame $I^t$}
\KwOut{\\
- Target State for new frame $X^t$\\
- Updated template buffer and reference frame}
Take set $\mathcal{T}$ of $l$ evenly spaced templates from the buffer\\
$P\leftarrow$ Convert $\mathcal{T}$ to point set\\ 
\For(\tcp*[f]{Forward pass}){every particle $X_i$ }{
	$C_i\leftarrow imcrop(I^{t},X_i)$\\
	$Q\leftarrow$ convert $C_i$ to point set\\
	$bbs_i = BBS(P,Q)$ \tcp*[f]{Random sampling!}\\
	$\pi_i = exp(bbs_i)$
}
$\pi_i = \frac{\pi_i}{\Sigma_i\pi_i}$\tcp*[f]{Normalize weights}\\
$X^t = \argmax\limits_{X} \pi_i$\tcp*[f]{Take MAP state}\\
$T^t = imcrop(I^t,X^t)$\tcp*[f]{Crop target template}\\
Perform backward pass according to Alg. \ref{alg:FWBK}\\
Check if template can be updated (Eq. \ref{eq:t-update})\\
Check if reference frame can be updated (Eq. \ref{eq:ref-update})\\
Use $\pi_i$ to draw new particle according to \cite{IsardB98}.
	\normalsize
\caption{Best-Buddies Tracker}
\label{alg:BBT}
\end{algorithm} 

\minesubsec{BBT Tracker Ensemble:} 
Setting the correct scale factor is critical when using particle filtering. A low scale factor will make the tracker conservative and more stable. However, it will not be able to cope with large and abrupt scale changes. A large scale factor on the other hand, can handle large scale changes, but is harder to control making the tracker less stable.

In order to cope with this problem we use an ensemble of several tracker with different  scale settings. Each such tracker is independent and does not exchange data with the other trackers. Tracker predictions are fused using the Online Trajectory Optimization method of Bailer et al.\cite{bailer2014superior}.

The final BBT ensemble tracker is summarized in Algorithm \ref{alg:ensemble}.
\begin{algorithm}[htb!]
\LinesNumbered
\small
\KwIn{New frame $I^t$}
\KwOut{\\
- Final target state $X^{fin}$\\
- Updated tracker states}
\For{every tracker $TRK_i$ in ensemble}{
	Perform tracking according to Alg. \ref{alg:BBT}\\
	Keep $X^t_{TRK_i}$\\
}
Fuse tracker predictions using the Online Trajectory Optimization\cite{bailer2014superior} to obtain $X^{fin}$.
	\normalsize
\caption{Best-Buddies Tracker Ensemble}
\label{alg:ensemble}
\end{algorithm} 
\section{Experimental Results}
We evaluate the performance of BBT on three commonly used data sets. (i) Object tracking benchmark 50, OTB-50~\cite{Wu12}, containing 50 sequences. (ii) Object tracking benchmark 100, OTB-100~\cite{wu2015object} containing 100 sequences (50 from OTB-50 and 50 additional sequences). (iii) Princeton tracking benchmark, PTB\cite{song2012tracking} which contains 95 sequences (this dataset also provides depth data however we only evaluate using RGB).

The performance of BBT is compared to other recently published tracking algorithms. Specifically, performance on OTB-50 and OTB-100 is compared with the following trackers: HCF~\cite{ma2015hierarchical}, MEEM~\cite{zhang2014meem}, DSST~\cite{DSST2015}, KCF~\cite{henriques2015high}, Struck~\cite{HareST11}, TGPR~\cite{TGPR2014}, SCM~\cite{SCM12}, STC~\cite{zhang2014fast}, PCOM~\cite{wang2015fast}. The performance on PTB is compared with 3D-T\cite{bibi20163d}, ASKCF~\cite{camplani2015real}, TGPR~\cite{TGPR2014}, KCF~\cite{henriques2015high}, Struck~\cite{HareST11}, VTD~\cite{KwonL10}, RGB~\cite{song2012tracking}, MIL~\cite{babenko2009visual}, TLD~\cite{KalalMM12},CT~\cite{zhang2012real}.

Performance is measured according to the OPE protocol (\cite{Wu12}) and is based on the intersection over union (IOU) criterion which quantifies both position as well as scale accuracy. A success curve measuring tracker success for accuracies ranging from 0 to 1 is built averaging over all the sequences in each data set. The mean average precision (mAP) is taken to be the area-under-curve (AUC) of the final overall success curve.

All our experiments are conducted with fixed parameters. We use an ensemble of 4 trackers with scale parameters $\{0,0,0.01,0.03\}$. Other parameters are fixed across trackers. Template buffer size is $L=30$, and $l=5$ templates are used for BBS computations. Template and reference frame update parameters are set to  $\gamma_1 = 0.6, f_1=5, \gamma_2=0.5 ,f_2=9$. We use 200 particles in each tracker and sample up to 300 points in the random sampling process described in Section \ref{sec:BBS sampling}. Using this configuration our unoptimized Matlab code runs at around 1 fps.

\begin{figure}
 	\centering
 	\begin{tabular}{c}
 		\includegraphics[width = 0.49\textwidth]{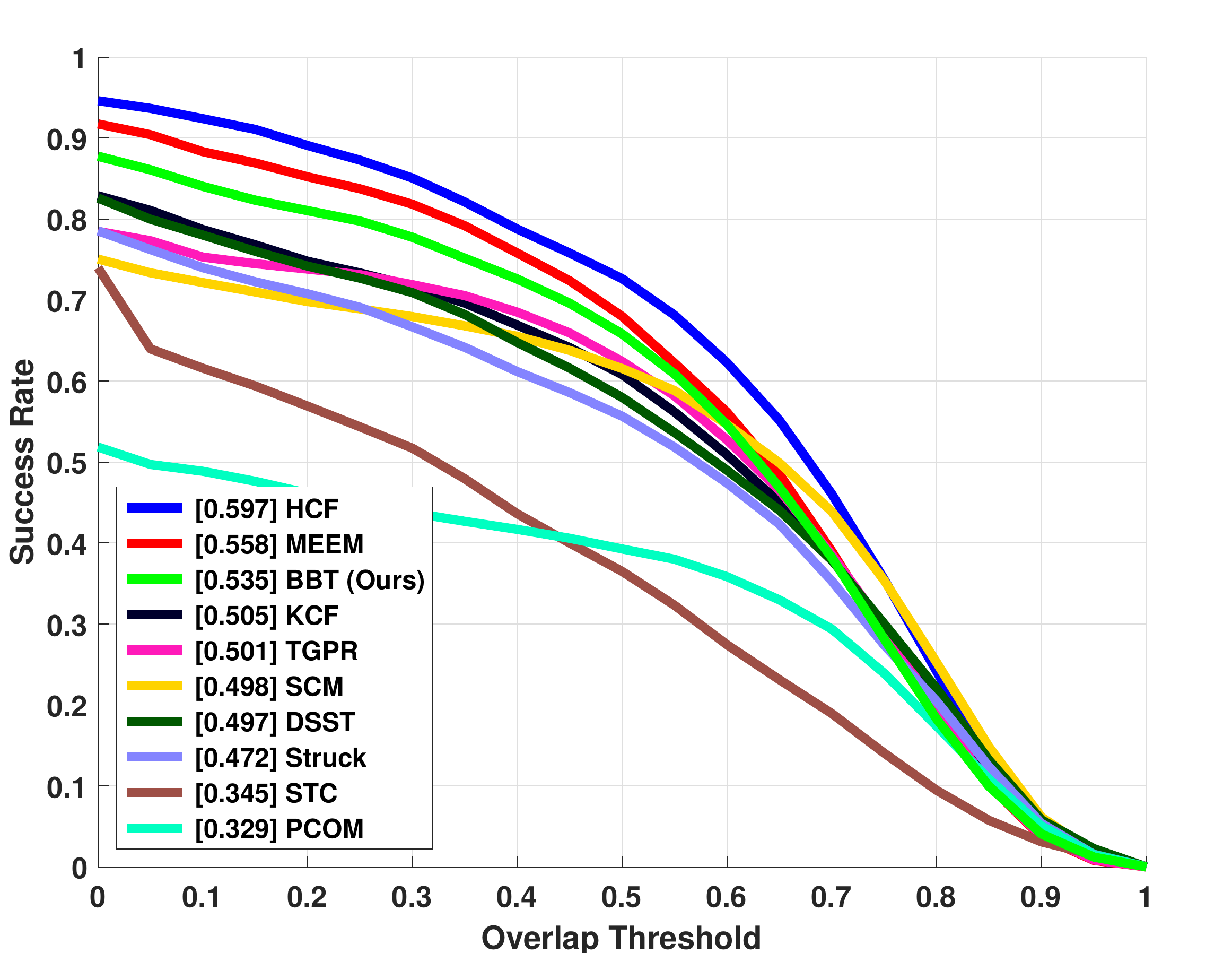} \\
 	\end{tabular}
 	\caption{{\bf Success plot for OTB-50\cite{Wu12}}. AUC shown in legend. BBT, the proposed method, shown in Green. Best viewed in color. }
 	\label{fig: SR OTB50} \vspace{-0.3cm}
 \end{figure}

\begin{figure}
 	\centering
 	\begin{tabular}{c}
 		\includegraphics[width = 0.49\textwidth]{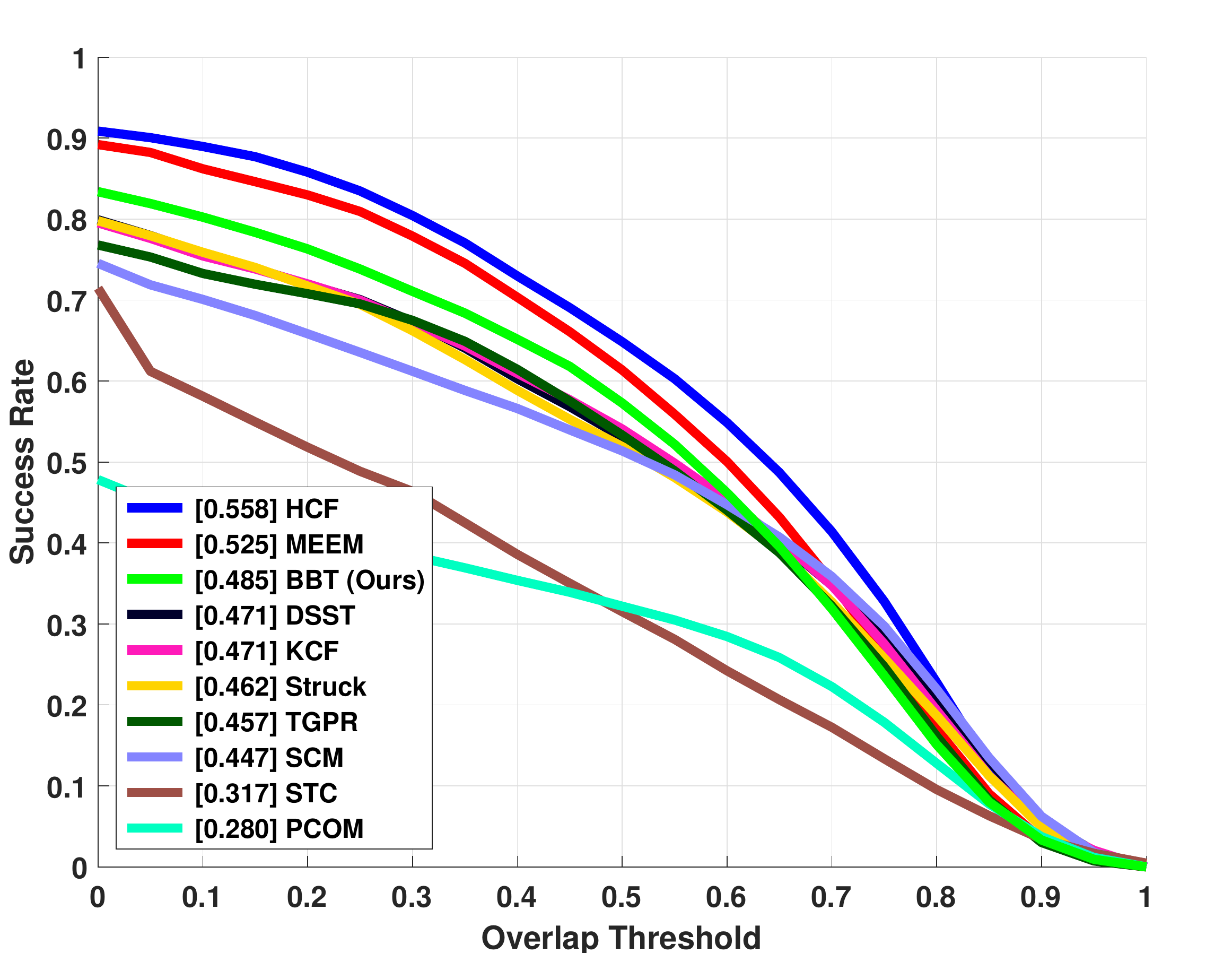} \\
 	\end{tabular}
 	\caption{{\bf Success plot for OTB-100\cite{wu2015object}}.AUC shown in legend. BBT, the proposed method, shown in Green. Best viewed in color.}
 	\label{fig: SR OTB100} \vspace{-0.3cm}
 \end{figure}

Results for OTB-50 and OTB-100 are presented in Figure \ref{fig: SR OTB50} and \ref{fig: SR OTB100} respectively. In OTB-50 (Fig. \ref{fig: SR OTB50}) , BBT comes in third place among the trackers evaluated with AUC of 0.535. It is able to outperform trackers such as KCF, DSST and STC, but is outperformed by HCF and MEEM. In OTB-100 (Fig.  \ref{fig: SR OTB100}) all tracker performances decrease. Overall BBT remains in third place with similar margins relative to competing trackers.

Results for the PTB are summarized in Table \ref{tbl:PTB}. As can be seen BBT comes in second place after 3D-T. Again BBT is able to outperform TGPR, KCF and additional recently published methods. Overall its performance in different categories is consistent with other tracking methods producing better results in easier scenarios i.e. rigid objects, slow motion etc. 
\begin{table*}[htb!]
\setlength{\tabcolsep}{2pt}     
\renewcommand{\arraystretch}{1} 
\centering
\caption{{\bf Tracking results for Princeton Tracking Benchmark\cite{song2012tracking}.} Success rate($\%$) and rankings (in parentheses) for different categories. BBT, the proposed method (in bold), is in the overall second place. }
\label{tbl:PTB}
\begin{tabular}{|l|c|c|c|c|c|c|c|c|c|c|c|c|}
\hline
\multirow{2}{*}{Alg.} & \multirow{2}{*}{\begin{tabular}[c]{@{}c@{}}Avg. \\ Rank\end{tabular}} & \multicolumn{3}{c|}{target type}                       & \multicolumn{2}{c|}{target size}    & \multicolumn{2}{c|}{movement}       & \multicolumn{2}{c|}{occlusion}      & \multicolumn{2}{c|}{motion type}    \\ \cline{3-13} 
                      &                                                                       & human            & animal           & rigid            & large            & small            & slow             & fast             & yes              & no               & passive          & active           \\ \hline
3D-T\cite{bibi20163d}\        & 1.09                                                                  & 0.81(1)          & 0.64(1)          & 0.73(1)          & 0.80(1)          & 0.71(1)          & 0.75(2)          & 0.75(1)          & 0.73(1)          & 0.78(1)          & 0.79(1)          & 0.73(1)          \\ \hline
\textbf{BBT (Our)}          	& \textbf{2.45}                                                         & \textbf{0.52(3)} & \textbf{0.55(2)} & \textbf{0.69(3)} & \textbf{0.58(3)} & \textbf{0.60(2)} & \textbf{0.77(1)} & \textbf{0.52(3)} & \textbf{0.52(3)} & \textbf{0.69(2)} & \textbf{0.70(3)} & \textbf{0.55(2)} \\ \hline
ASKCF\cite{camplani2015real}  & 2.64                                                                  & 0.52(2)          & 0.50(4)          & 0.72(2)          & 0.59(2)          & 0.59(3)          & 0.67(3)          & 0.56(2)          & 0.52(2)          & 0.68(4)          & 0.72(2)          & 0.54(3)          \\ \hline
TGPR\footnote{Results were not formally submitted and, are only available as raw data, therefore rank is not provided.}
\cite{TGPR2014}& N/A                                                                   & 0.46             & 0.49             & 0.67             & 0.56             & 0.53             & 0.66             & 0.5              & 0.44             & 0.69             & 0.67             & 0.5              \\ \hline
KCF\cite{henriques2015high}   & 3.82                                                                  & 0.42(4)          & 0.50(3)          & 0.65(4)          & 0.48(4)          & 0.55(4)          & 0.65(4)          & 0.47(4)          & 0.41(4)          & 0.68(3)          & 0.65(4)          & 0.47(4)          \\ \hline
Struck\cite{HareST11}         & 5.91                                                                  & 0.35(5)          & 0.47(7)          & 0.53(7)          & 0.45(5)          & 0.44(7)          & 0.58(5)          & 0.39(5)          & 0.30(7)          & 0.64(5)          & 0.54(7)          & 0.41(5)          \\ \hline
VTD\cite{KwonL10}             & 6.09                                                                  & 0.31(7)          & 0.49(5)          & 0.54(6)          & 0.39(6)          & 0.46(5)          & 0.57(6)          & 0.37(6)          & 0.28(8)          & 0.63(6)          & 0.55(6)          & 0.38(6)          \\ \hline
RGB\cite{song2012tracking}    & 7.27                                                                  & 0.27(10)         & 0.41(8)          & 0.55(5)          & 0.32(10)         & 0.46(6)          & 0.51(8)          & 0.36(7)          & 0.35(5)          & 0.47(9)          & 0.56(5)          & 0.34(7)          \\ \hline
MIL\cite{babenko2009visual}   & 8.64                                                                  & 0.32(6)          & 0.37(9)          & 0.38(9)          & 0.37(8)          & 0.35(9)          & 0.46(10)         & 0.31(8)          & 0.26(9)          & 0.49(8)          & 0.40(11)         & 0.34(8)          \\ \hline
TLD\cite{KalalMM12}           & 8.64                                                                  & 0.29(9)          & 0.35(10)         & 0.44(8)          & 0.32(9)          & 0.38(8)          & 0.52(7)          & 0.30(10)         & 0.34(6)          & 0.39(10)         & 0.50(8)          & 0.31(10)         \\ \hline
CT\cite{zhang2012real}        & 8.73                                                                  & 0.31(8)          & 0.47(6)          & 0.37(10)         & 0.39(7)          & 0.34(10)         & 0.49(9)          & 0.31(9)          & 0.23(11)         & 0.54(7)          & 0.42(10)         & 0.34(9)          \\ \hline
\end{tabular}
\vspace*{-2mm}
\end{table*}

We note that we find the fusion step to be a limiting factor on performance. For example, analyzing our performance on OTB-50 reveals that if we were able to choose the optimal tracker per \emph{sequence} (choose the correct scale factor) we could reach mAP of 0.571. Furthermore, choosing the best tracker per \emph{frame} would result in mAP of 0.641. In light of these findings, as part of our future research, we plan to search for a better fusion techniques delivering better overall performance.   
\section{Conclusions}
The Best-Buddies Similarity between point sets, has been successfully applied to template matching, showcasing an ability to handle non-rigid deformations and automatically reject outliers, making it attractive for visual tracking.


Applying BBS to tracking requires it to handle point sets of arbitrary size in order to cope with things such as scale changes and using multiple templates. In this work we found BBS to be biased when computed between point sets with different sizes. A theoretical as well as empirical study of this problem lead us to two effective solutions: clustering and random sampling. We found random sampling favorable and more accurate as it requires no preprocessing, has no associated weights, does not alter the underlying distribution of the data and can be computed in constant time. 


Using random sampling we were able to successfully apply BBS to visual tracking. This was done by integrating BBS into a particle filtering framework. By augmenting data from multiple templates we were able to extend BBS to handle the temporally varying appearance of objects being tracked, and an ensemble of BBT tracker was used to ensure good scale estimation.
 
Extensive experiments were performed using three commonly used tracking benchmarks. BBS demonstrated good initial performance, competitive with respect to other recently published tracking algorithms. 

One of the main limiting factors on performance was found to be the fusion technique used. In light of this, our future research is aimed at finding a better fusion strategy that can lead to better performance.
\section*{Appendix - Proof of Claim from Section \ref{sec:bbs_claim}}
\minesubsec{}
Let the minimal distance between $p_i$ and any other point in $P$ be,
\begin{equation}
	2\varepsilon = \min_{p_j\in P , i\neq j} d(p_i,p_j) = \min_{p_j\in P , i\neq j} d_{ij}
\end{equation}
Where we use $d_{ij}$ as a shorthand notation for $d(p_i,p_j)$ which is the distance between points $p_i$ and $p_j$. We note that since we are dealing with continuous distributions and since $|P|$ is finite then $\forall i \neq j\quad p_i\neq p_j$ and therefore $\varepsilon > 0$.

By construction, $p_i$ is the nearest neighbor of any point $q_j \in Q$ such that $d_{ij}<\varepsilon$ in other words, if $d_{ij}<\varepsilon$ then $NN(q_j,P)=p_i$. If there is exactly one such point then $NN(p_i,Q)=q_j$ and then $bb(p=p_i,q,P,Q) = 1$. It is easy to see that, if there is more than one point $q\in Q$ that satisfies $d_{ij}<\varepsilon$. Then, since $p_i$ is the nearest neighbor to all of these points, one of them will be the nearest neighbor of $p_i$ in $Q$ and again $bb(p=p_i,q,P,Q) = 1$.

This means that if there is at least one point $q_j \in Q$ such that $d_{ij}<\varepsilon$ then $p_i$ will have a BBP. Formally, we want to show that,
\begin{equation}\label{cl1}
	pr(\exists q_j \in Q \text{ s.t. } d_{ij}<\varepsilon) \underset{M\rightarrow+\infty}{\rightarrow} 1
\end{equation}
Similarly we can check what is the probability that no point in $Q$ is $\varepsilon$-close to $p_i$,
\begin{equation}	
	pr(\exists q_j \in Q \text{ s.t. } d_{ij}<\varepsilon) =\\ 1 - pr(\forall q_j \in Q \text{ , } d_{ij}\geq\varepsilon) 
\end{equation}
The probability that a point $q_j$ was randomly drawn $\varepsilon$-close to $p_i$, requires integrating $f_Q(q)$ over a $\varepsilon$-hypersphere around $p_i$. We denote this integration region as $\Phi_{p_i,\varepsilon}$, and then we have
\begin{equation}\label{eq_1}
	pr(q_j \in \Phi_{p_i,\varepsilon}) =  \underset{\Phi_{p_i,\varepsilon}}{\int\int\dotsb\int}f_Q(p)dp
\end{equation}
Note this is a multivariate distribution, and we are integrating over all the dimensions. 

Solving this integral for some arbitrary distribution can be very difficult. Fortunately, we are only interested in bounding it. Specifically, since $f_Q$ is a smooth distribution function and since $\varepsilon>0$ then,
\begin{equation}\label{eq_2}
	1 > pr(q_j \in \Phi_{p_i,\varepsilon}) > 0
\end{equation}
The probability that a point will \emph{not} be $\varepsilon$-close to $p_i$, i.e. $d_{ij}\geq\varepsilon$ , is given by the compliment probability of equation (\ref{eq_1}). Since the points are i.i.d. we can factor over all the points and get the probability that \emph{all} points in $Q$ are \emph{not} $\varepsilon$-close to $p_i$:
\begin{equation}
	pr(\forall q_j \in Q \text{ , } d_{ij}\geq\varepsilon)  = [1-pr(q_j \in \Phi_{p_i,\varepsilon})]^M
\end{equation}
Using the bounds in (\ref{eq_2}) we have that the base of this power is smaller than one, and since the power $M\rightarrow\infty$ we have that,
\begin{equation}
	pr(\forall q_j \in Q \text{ , } d_{ij}\geq\varepsilon  )\underset{M\rightarrow+\infty}{\rightarrow} 0 
\end{equation}
Which means the limit in (\ref{cl1}) holds, and thus,
\begin{equation}
	\forall p_i\in P \text{ , } pr(bb(p=p_i,q,P,Q) = 1) \underset{M\rightarrow+\infty}{\rightarrow} 1
\end{equation}
That is, all point in $P$ find a BBP, and since we normalize by the size of $P$, which is the smaller set, the BBS score goes to 1, and we are done.

{\small
\bibliographystyle{ieee}
\bibliography{BBT_bib}
}

\end{document}